\documentclass[5p]{elsarticle}
\usepackage{natbib}
\usepackage{amsmath,amssymb}
\usepackage{graphicx}
\usepackage{textcomp}
\usepackage{xcolor}
\usepackage{color}
\usepackage{multicol}
\usepackage{color}
\usepackage{subfigure}
\usepackage[english]{babel}
\def\BibTeX{{\rm B\kern-.05em{\sc i\kern-.025em b}\kern-.08em
    T\kern-.1667em\lower.7ex\hbox{E}\kern-.125emX}}

\begin{document}

\title{Deep learning using Havrda-Charvat entropy for classification of pulmonary optical endomicroscopy\\}

\author{Thibaud BROCHET}
\address{LITIS, Eq. Quantif, University of Rouen}

\author{J\'er\^{o}me LAPUYADE-LAHORGUE}
\address{LITIS, Eq. Quantif, University of Rouen}

\author{S\'ebastien BOUGLEUX}
\address{GREYC, Eq. Image, Ensicaen}

\author{Mathieu SALA\"UN}
\address{University Hospital of Rouen}

\author{Su RUAN}
\address{LITIS, Eq. Quantif, University of Rouen}

\begin{abstract}
Pulmonary optical endomicroscopy (POE) is an imaging technology in real time. It allows to examine pulmonary alveoli at a microscopic level. Acquired in clinical settings, a POE image sequence can have as much as $25\%$ of the sequence being uninformative frames (i.e. pure-noise and motion artefacts). For future data analysis, these uninformative frames must be first removed from the sequence. Therefore, the objective of our work is to develop an automatic detection method of uninformative images in endomicroscopy images. We propose to
take the detection problem as a classification one. Considering advantages of deep learning methods, a classifier based on CNN (Convolutional Neural Network) is designed with a new loss function based on Havrda-Charvat entropy which is a parametrical generalization of the Shannon entropy. We propose to use this formula to get a better hold on all sorts of data since it provides a model more stable than the Shannon entropy. Our method is tested on one POE dataset including 2947 distinct images, is showing better results than using Shannon entropy and behaves better with regard to the problem of overfitting.
\end{abstract}


\maketitle
\pagebreak
\section{Introduction}
\label{sec:intro}
\begin{figure}[h!]
\begin{center}
\subfigure{\includegraphics[width=0.1\textwidth]{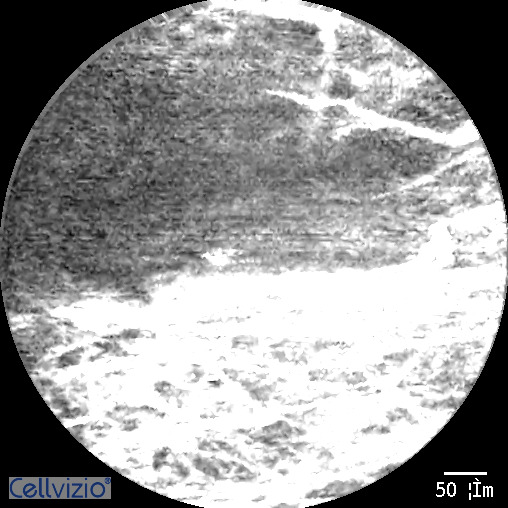}} \subfigure{\includegraphics[width=0.1\textwidth]{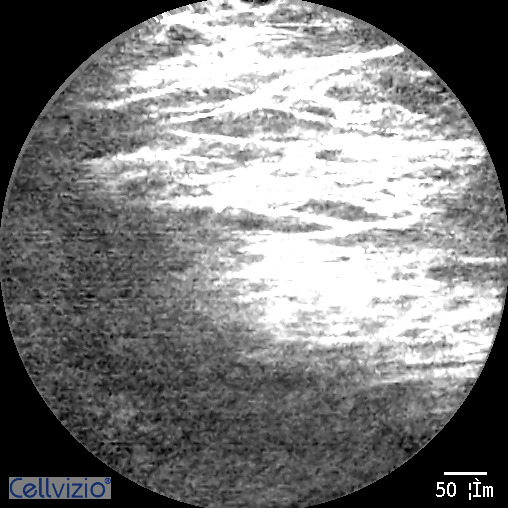}} \subfigure{\includegraphics[width=0.1\textwidth]{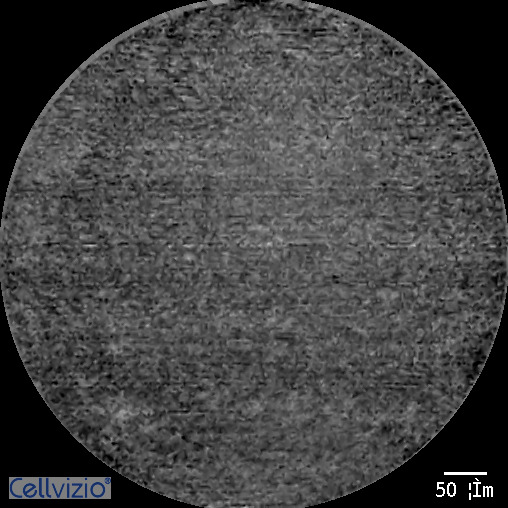}} \subfigure{\includegraphics[width=0.1\textwidth]{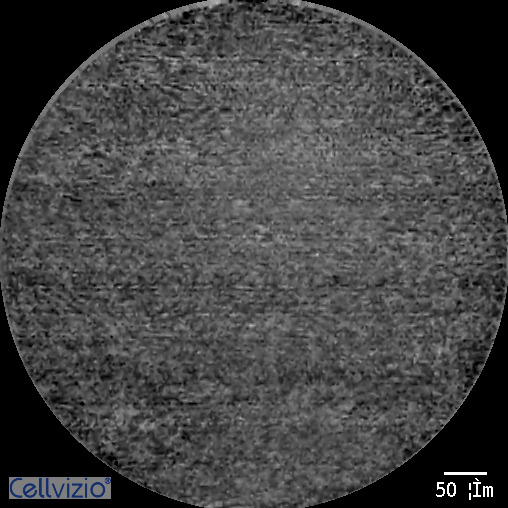}}\\
\subfigure{\includegraphics[width=0.1\textwidth]{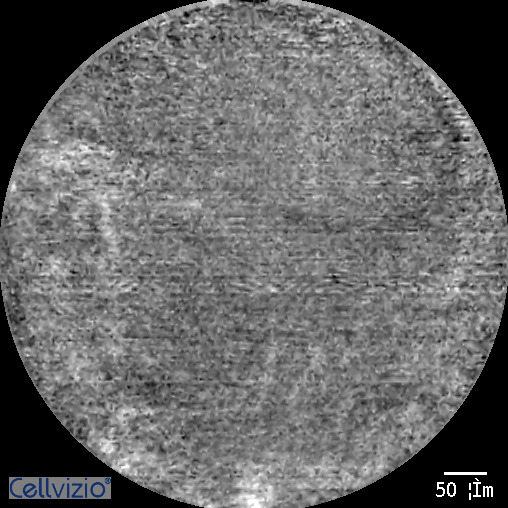}} \subfigure{\includegraphics[width=0.1\textwidth]{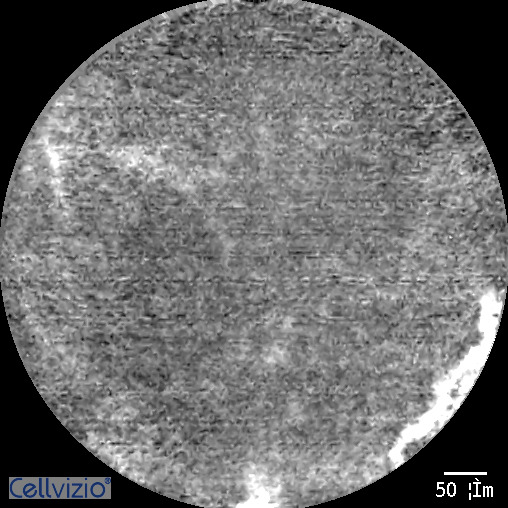}} \subfigure{\includegraphics[width=0.1\textwidth]{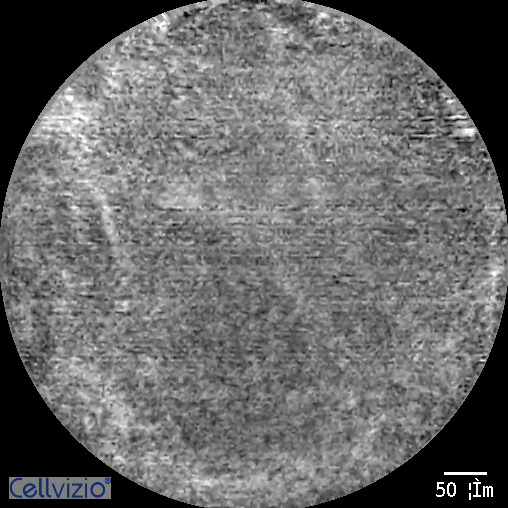}} \subfigure{\includegraphics[width=0.1\textwidth]{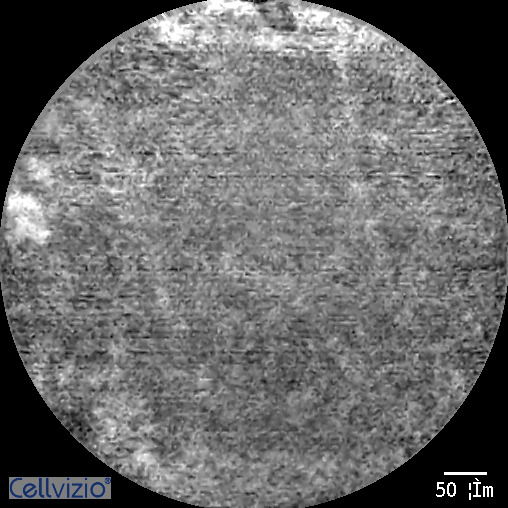}}\\
\caption{\label{fig:informative_vs_uninformative} A sequence of images from a patient. First row: two informative, then two uninformative frames. Second row: two uninformative frames then two informative ones}
\end{center}
\end{figure}

Pulmonary optical endomicroscopy uses fiber confocal fluorescence microscopy which can provide diagnostic information about fibrosis and inflammation of the distal air spaces associated with lung disease \cite{salaun2012}. It is a new, real-time imaging technology that provides pulmonary alveoli imagery at a microscopic level. However, acquired in clinical use, a POE image sequence can have a proportion of more than $25\%$ of the sequence giving uninformative frames as pure-noise and motion artefacts \cite{perperidis2017}. For a future data analysis, these uninformative frames must be first removed from the dataset. In clinical examination, the detection of uninformative frames is actually carried out manually. This manual operation is time consuming and laborious. Therefore, automatic detection is necessary to speed up data analysis and shorten diagnostic time. Our work aims at developing an automatic detection method to remove uninformative frames from a sequence of images. Fig. \ref{fig:informative_vs_uninformative} shows four informative and four non-informative images. We can observe that textures in these two kinds of images are very different. Hence, we can consider the detection problem as a classification one.

Deep learning is an advanced machine learning technique, showing powerful classification ability. It consists in estimating the parameters of the activation functions by minimizing a loss function comparing the output of the neural network and the desired output. It has shown to be outperforming the state-of-the art specialists \cite{Zhou2019,Li2020} in various applications. One class of deep learning techniques is the supervised deep learning techniques. In supervised deep learning techniques, the considered data are labeled; for instance: informative versus uninformative POE. The loss function evaluates the difference between the estimated labels from the network and the true labels. Parameters are updated in order to minimize the loss. To achieve this, we generally use a Convolutional Neural Network (CNN) \cite{fukushima1988,lecun1998}. Fig. \ref{fig:architecture} presents the general architecture of CNNs. 

 Many methods of classification based on CNN have been developed in medical image fields \cite{yadav2019,ahn2016,heung2017,amyar2019}. Definition of the loss function for training the network is important for achieving good performance of accuracy. Most studies using entropy in deep learning were focused on most common forms as quadratic metric or Shannon cross-entropy. Quadratic metric is generally used when the desired ouput is deterministic; for instance in image segmentation. Cross-entropy loss functions are generally used for probabilistic decision; for instance in classification. Shannon entropy is the most known entropy and the derived cross-entropy can be easily interpreted as a ``distance'' between probabilities \cite{Sen2015}. Moreover, Shannon entropy is more relevant when the data follows a Gibbsian distribution (ie. exponential of the opposite of a convex energy function). Unlike Shannon's entropy, Havrda-Charvat entropy \cite{Roselin2014,Chen2010,Maksa2008,Kurt2014,Zhu2020} doesn't require a specific distribution of data in order to keep its specificity and accuracy. 
In this paper, we propose to implement a CNN with a loss function derived from Havrda-Charvat's entropy to classify the studied frames in their proper categories (informative/uninformative) and to compare the results with a Shannon-based cross-entropy and find out what improvement can be achieved.
\section{Related Works}
\subsection{Data classification}
Their exist many methods for classifying data. Some methods as Support Vector Machines (SVM) are geometrical \cite{cortez1995}. SVM is inspirated from the Hahn-Banach theorem \cite{brezis} and is a supervised learning method consisting in separating two convex subsets by an hyperplan. SVM has been generalized in non-linear case by the way of the introduction of a metric kernel \cite{Crammer2001} and \cite{sollich2002} proposes to estimate the kernel via Bayesian inference. The decision tree method \cite{breiman1984} constructs classifiers by tree both in regression and in classification. So a tree is built by gradually dividing a population into two sub-populations in order to optimize the homogeneity of populations according to their label. Decision trees have been generalized into random forests \cite{breiman2001} which group together a multitude of independent decision trees. These are built from the same learning base using different random processes. The fact of combining several decision trees makes it possible to reduce the influence of noisy data during the learning phase. Random forests have been used successfully in classification; for instance in radiomics \cite{parmar2015}.  Other classification methods are statistical. Amongst statistical methods, logistic regression \cite{vittinghoff2006} consists in learning parameters of the logistic classification function from annoted data. In likelihood maximization \cite{barbu2006}, a probabilistic model representing the probability of each class in function of the observation is given; the decision consists in choosing the most likely class. Such likelihood maximization is also used in the more recent Conditional Random Fields (CRF) for image segmentation \cite{magnano2014} and for classification \cite{liliana2017}. Another kind of statistical methods are the Bayesian methods in which we have a prior knowledge about the belonging to a class. Bayesian methods \cite{bernardoSmith1994,ghosh2006} have been widely used in image segmentation \cite{lanchantin2008} and data classification \cite{priya2015}. More recently, the deep-learning methods is more and more used. Amongst applications of deep learning in data classification, one can quote text recognition \cite{audebert2019}, face recognition \cite{majeedi2018} or spams detection \cite{dada2019}. The most classical deep-learning method for supervised classification is based on Convolutional Neural Networks (CNN). Generally, a CNN is composed of two sets of layers. The first one applies convolution maps in order to reduce the data and the second one is a fully connected network. There are different CNN-based architectures. Amongst them, the LeNet architecture \cite{lecun1998} is the first successful CNN used for classify digits; AlexNet \cite{krizhevsky2012} was the first CNN applied to computer vision and was submitted to the ImageNet ILSVRC challenge in 2012; ZFNet is an improvement of AlexNet proposed in \cite{zeiler2013}. Many applications of deep learning for classification of optical images have already been made in \cite{moccia2020,li2019,chang2018}.
\subsection{Generalized entropies and its application}
In a technical point of view, there are several ways to generalize the classical Shannon entropy and the different metrics and divergences as summarized in \cite{basseville2010}. There is two main ways to generalize the Shannon entropy: the first one consists in replacing the integrated functional in the expression of the Shannon entropy and the second one according to an axiomatic \cite{kumar2014}. As underlined in \cite{Li2020}, Havrda-Charvat and Shannon entropies are the only ones which satisfy the strong recursivity property. Even if generalization of Shannon entropy as Havrda-Charvat is not recent; most applications of them is clustering \cite{Li2020,Sen2015,Chen2010,Zhu2020} or coding theory \cite{kumar2012InternalJournal}. In \cite{kumar2014}, a weakened recursivity property is studied and generalization of the Havrda-Charvat satisfying this property is proposed. This property makes easier it use in case of multi-label classification as, for instance, for gene expression analysis \cite{Li2020}. In \cite{Roselin2014}, results of mammogram image classification are compared while using different parameterized entropies: R\'enyi, Havrda-Chravat, Tsallis and Kapur's entropy. It appears that Tsallis entropy gives the best results as part of their study. However, Tsallis entropy has two parameters whereas Havrda-Charvat has only one parameter and gives slighly less performant results. Havrda-Charvat entropy has also been used for clustering in \cite{Chen2010} by replacing the Shannon entropy by the Havrda-Charvat entropy in the Jensen-Shannon divergence. In \cite{Maksa2008}, a functional equation from which Havrda-Charvat, Shannon and Tsallis entropies is proposed. Moreover, the stability in the sense of Hyers-Ulam \cite{hyers2004} of this functional equation is studied.

\section{Method}
\label{sec:methodology}
\begin{figure}[!h]
\begin{center}
\includegraphics[height=4cm,width=9cm]{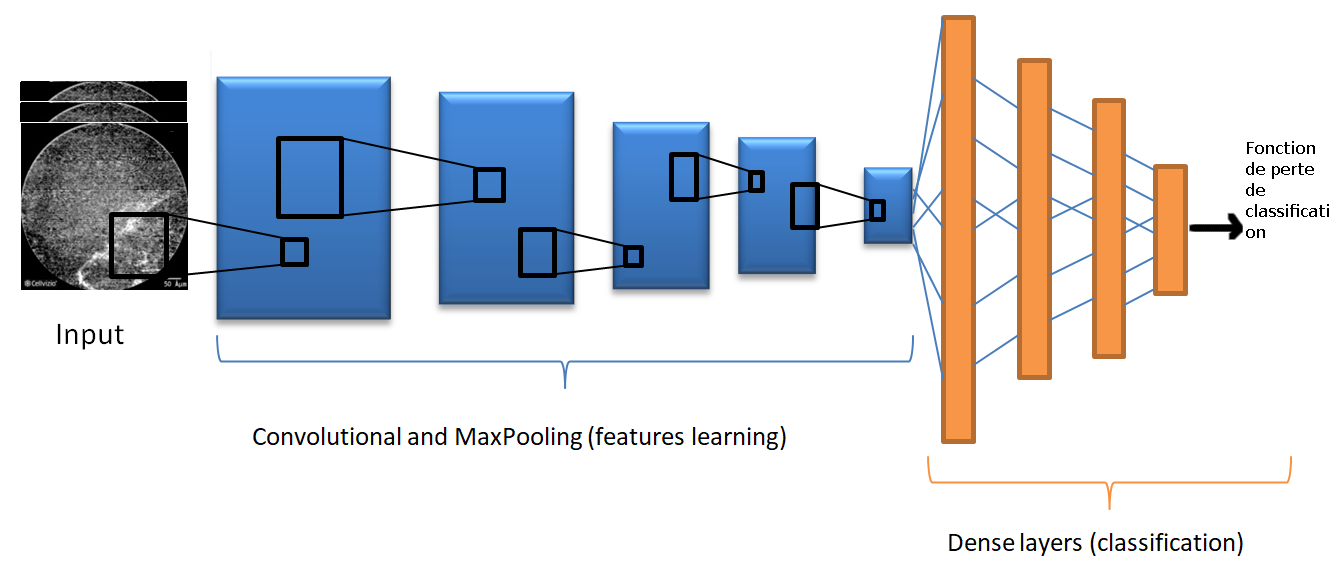}
\caption{\label{fig:architecture} Architecture of the CNN for supervised learning}
\end{center}
\end{figure}

In this section, we present the neural network that we use for the classification of the optical images.\\
We use a supervised method for the classification of optical images of lungs into two classes: informative class and non-informative  class. The method that we use is a Convolutional Neural Network (CNN) whose architecture is represented in Fig. \ref{fig:architecture}.  The loss functions we use for the supervised learning are entropy-based functions. These loss functions compare two probability laws. The first one is the output of the CNN and represents the estimated probability of the input image belonging to informative class. This estimated probability is computed by the sigmoid stage of the CNN. The second probability is a Dirac probability whose value is $1$ if the image is informative one and $0$ otherwise. A entropy-based loss function is constructed by choosing an entropy function and a divergence for comparing the entropies of the two distributions. The first entropy is the classical Shannon entropy and the second one is the Havrda-Charvat entropy which generalizes the Shannon entropy.
\subsection{Supervised classification of optical images}
In this paper, we focus on classifying optical images into two classes informative and non informative images. The set of possible states is $\Omega=\left\{0,1\right\}$ where $1$ corresponds to the event ``informative image'' and $0$ to ``non informative image''. In supervised classification, we train the network from an annotated database. Let $N$ be the number of annotated images. At the $n$-th image, we associate the following Dirac probability:
\begin{equation}
q_{n}(1)=\left\{\begin{array}{c}
1\textrm{ if the image is informative}\\
0\textrm{ otherwise}
\end{array}\right.
\end{equation}
Let $p_n$ the output of the CNN for the $n$-th image. The loss function to be minimized is:
\begin{equation}
L(q,p)=\sum_{n=1}^{N}H(q_n,p_n),
\end{equation}
where $H$ is the chosen entropy-divergence.\\
The two following paragraphs detail how one can generalize the Shannon entropy and how an entropy-divergence can be built from an entropy.
\subsection{Generalized entropies}
An entropy function is a concave function from a subset of probability densities to the real line. The most known entropy function is the Shannon entropy, for discrete probabilities (i.e. probabilities defined on a countable space $\Omega$, this one is defined by:
\begin{equation}
H_{\nu}(p)=-\sum_{\omega\in\Omega}\log(p(\omega))p(\omega)\nu(\omega),
\end{equation}
where $\nu$ is a measure on  $\Omega$; in practice, $\nu$ is taken to be the counting measure but can also represents a prior information on the state $\omega$.\\
As explained in \cite{basseville2010}, there are several ways to generalize the Shannon entropy. A classical way to define generalized entropy is:
\begin{equation}
\label{eq:generalized_entropy}
H_{\nu}(p)=-\sum_{\omega\in\Omega} h(p(\omega))\nu(\omega)
\end{equation}
where $h$ is a convex function defined on $\mathbb{R}$.\\
In this paper, we propose to utilise the Havrda-Charvat entropy \cite{kumar2014} which belongs to a parametric family whose convex functional is given by:
\begin{equation}
h_{\alpha}(p)=\frac{p^{\alpha}-p}{\alpha-1},
\end{equation}
where $\alpha>1$. By studying the limit at $\alpha=1$, we find $h_{1}(p)=p\log(p)$ back. As a consequence, the Havrda-Charvat whose parameter is equal to $1$ coincides with the Shannon entropy.\\
By replacing $h$ by $h_{\alpha}$ in (\ref{eq:generalized_entropy}), we deduce the expression of the Havrda-Charvat entropy:
\begin{equation}
\label{eq:havrda-charvat}
H_{\alpha,\nu}=\frac{1}{\alpha-1}\sum_{\omega\in\Omega}\left[p(\omega)-p(\omega)^{\alpha}\right]\nu(\omega)
\end{equation}
and if $\nu$ is the counting measure:
\begin{equation}
\label{eq:havrda-charvat-bis}
H_{\alpha,\nu}=\frac{1}{\alpha-1}\left(1-\sum_{\omega\in\Omega}p(\omega)^{\alpha}\right)
\end{equation}
\subsection{Entropy-divergences and cross-entropy}
There exist several ways to construct a loss function from a given entropy.
In this paper, we consider the cross-entropy, as it is the most used in deep-learning when the entropy is the Shannon entropy. This one is defined as:
\begin{equation}
\label{eq:generalized_cross_entropy}
H(q,p)=-\sum_{\omega\in\Omega} h(p(\omega))\frac{q(\omega)}{p(\omega)}
\end{equation}
In the case where $h(p)=p\log(p)$, we can find the classical cross-entropy:
\begin{equation}
\label{eq:shannon_cross_entropy}
H_{1}(q,p)=-\sum_{\omega\in\Omega} q(\omega)\log(p(\omega))
\end{equation}
and for $h_{\alpha}(p)=\frac{p^{\alpha}-p}{\alpha-1}$, the Havrda-Charvat cross-entropy is defined by:
\begin{equation}
\label{eq:havrda_charvat_cross_entropy}
H_{\alpha}(q,p)=\frac{1}{\alpha-1}\times\left(1-\sum_{\omega\in\Omega}(p(\omega))^{\alpha-1}q(\omega)\right)
\end{equation}
In this paper, we compare classification results for several loss functions: Havrda-Charvat-based loss function with $\alpha$ ranging from 1 to 2.
\section{Results and analysis}

\subsection{Data}
In this study, the dataset consists of different kinds of datasets. The potential diseases are: asbestosis, pulmonary idiopathic fibrosis, hypersensitivity pneumopathy, sclerodermia and some healthy people as the control group. we have used 3895 images where 2313 were informative and 1582 were uninformative images. These frames were obtained from images sequences given by the CHU (Universitary Central Hospital) of Rouen, Normandy. The size of images are between 512x512 and 500x500 pixels, depending on the sequence. We proposed, to normalize all images in a 128x128 resolution, in order to train the CNN with a reasonable resolution by considering a balance between the image size and processing time. The data were obtained as video sequences and turned into frames for the analysis.

\subsection{Implementation}
We proposed to go for a simple CNN architecture. The two main kind of layers in a CNN are the following:
\begin{itemize}
\item 1) The first layers are the Convolutional layers. Their function is to reduce the input to its most prominent features, in order to retain the most important information while reducing the dimensions of the input image. These layers are often associated with an activation function, which is used to determine the value of the output, and eventually a MaxPooling, that aims to increase even further the reduction of dimensions and the extraction of features. 
\item 2) The second layers are Fully Connected layers, whose role is to establish a decision rule that leads, on the last layer, to the classification of the input image.
\end{itemize}
 We implemented the following layers: five convolutional layers , each having a convolution filter of dimensions (3,3) and using the activation method Rectified Linear Unit (ReLU) of decreasing size  (128, 64, 32, 16 and 8), and then four dense layers of decreasing size (128, 64, 32 and 16). The total number of trainable parameters in this architecture was around 6,6 millions. The validation split of this network was set at 70/30, i.e. that means 70\% served as a training set, and the remaining 30\% were used as a validation set in order for the CNN to improve. The batch size, the number of data simultaneously passed through the CNN, was 64. We also added Dropout layers after the first three Dense ones. These layers force the network to drop certain links, reducing the phenomenon of overfitting.

\subsection{Learning conditions}
We use  balanced datasets, with about half images being uninformative and half being informative.  This way we have prevented the availability bias that would've oriented the model toward the most available kind of data.
Such bias can potentially skewer data where nothing like that can be explained by data distribution.
We used the Python language with the Keras library which is a powerful easy-to-use Python library for developing and evaluating deep learning models.

The criteria of evaluation were those implemented with the CNN, i.e the amount of correct identification over the total length of the dataset.

\subsection{Experimentation}

A comparison study between different entropies is carried out in function of the number of epochs to define their accuracy (see TABLE 1). N represents the number of epochs set in the network's training, and $\alpha$ is the coefficient of identical name in the Havrda-Charvat's formula. The following table was made for a 2947 images dataset (1625 informative / 1322 uninformative):

\begin{table}[h!]
\begin{center}
  \begin{tabular}{|l|l|l|l|l|l|}
    \hline
    N& \multicolumn{5}{c|}{$\alpha$} \\\cline{2-6}

& 1.0 &1.1 & 1.3  & 1.5 &  2.0\\
    \hline

 20 & 0.59  &  0.65 & 0.62 & 0.63&0.59\\
\hline
 30 & 0.59  & 0.68  &0.73 &0.59 & 0.59\\
\hline
 40 & 0.70  &  0.79 &0.77 & 0.76&  0.59 \\
\hline

  \end{tabular}

\caption{\label{tab:results} Quantitative comparison of accuracy with different entropies. Sh.: Shannon, H-CH.: Havrda-Charvat}
\end{center}
\end{table}

To help assert our observations, we also computed the specificity and sensitivity for the same values of $\alpha$ for 40 epochs.

\begin{table}[h!]
\begin{center}
  \begin{tabular}{|l|l|l|l|l|l|l|}
    \hline

$\alpha$ & 1.0 &1.1 & 1.3  & 1.5 &  2.0\\
    \hline

 Specificity&  0.87 &0.91   &0.92  &0.78  &0.99\\
\hline
 Sensitivity & 0.71 & 0.79  & 0.78 & 0.75&  0.59\\
\hline

  \end{tabular}

\caption{\label{tab:results2} Comparison of sensitivity and specificity for several values of $\alpha$ with N=40 epochs}
\end{center}
\end{table}

We can observe that the number of epochs has moderate influence on the end results. Except, of course, when the number of epochs is deliberately too small to properly reach convergence.  During testing, we noticed that the results tended to present a greater disparity the fewer epoch we set. This is due to the fact that the network selects a random set of input images from the dataset to perform both training and testing. Then, by increasing the number of epoch, we make sure that the network reaches convergence and thus diminish the random factor in the subset's analysis.

The values of sensitivity and specificity allow to rule out the following value of $\alpha$: 2.0. These results concur with the previous ones. The remaining values are satisfactory regarding the model's accuracy.

When increasing the number of epochs beyond 40, we obtained very high results, including with Shannon's entropy. This can be explained by the fact that our sample remains small in comparison to what a Neural Network might need, and its learning can turn into overfitting, which is the network learning so well to recognize the patterns it has been given that it will be unable to recognize other ones, provided it differs too much from the training sets.
As a rule of thumbs, the more epoch the better, but if a network seems to have reached a stability after a certain number of epochs, the following ones tend to only improve them by a short amount, and that's when the network enters the domain of overfitting, which is not to be encouraged. That's why we need to carefully study how the network converges and prevent it from doing so for too long.

 The best results for the supervised elements were 79\% of correct classification when using Havrda-Charvat with alpha=1.1 (1.3) and 40 epochs. It's a slight, yet noticeable improvement from Shannon's entropy. Moreover, for 40 epochs, the respective results for Shannon's entropy and Havrda-Charvat($\alpha=1.1$) are 70\% and 79 \% of accuracy for the first experiment and 77\% and 79\% for Shannon and Havrda-Charvat($\alpha=1.3$) for the second experiment, which makes the later a huge improvement over Shannon. It demonstrates that Havrda-Charvat's formula can be an improvement when it comes to classification and the loss function that comes from it is a better alternative to the more common one.

 As a trend, we've noticed that the Havrda-Charvat formula yielded better results than Shannon's. It can be explained by the fact that Shannon's entropy needs data of a certain type to be totally relevant, i.e data that can be distributed on the exponential of a convex function, in other words it needs a single extremum to safely move toward it. If the data are not meeting this requirement, Shannon's entropy cannot be considered as completely reliable. Havrda-Charvat is a generalized version of Shannon's entropy, not needing any specific conditions for the data. The choice of the parameter $\alpha$ in Havrda-Charvat can be fixed depending on data or estimated.
Considering non any prior information about the nature of our data, this more generalized entropy can thus give better performance.

We can also observe that the  factor $\alpha$ is not to be increased to high values. When we got close from 2, quickly, after 20 epochs, we noticed that our network stopped improving and was stuck to accuracy values close to 0.60. \\
 The interval of $\alpha$ for good results is between 1.1 and 1.5 in both experiments. We will study how to automatically estimate this value in upcoming work.

\subsection{Validation loss study}

\begin{figure}[h!]
\begin{center}
\subfigure{\includegraphics[width=0.35\textwidth]{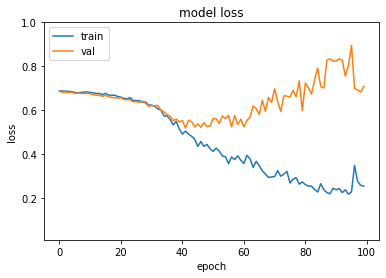}} \subfigure{\includegraphics[width=0.35\textwidth]{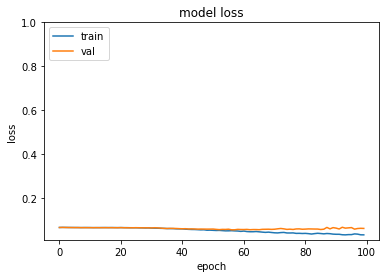}}  \subfigure{\includegraphics[width=0.35\textwidth]{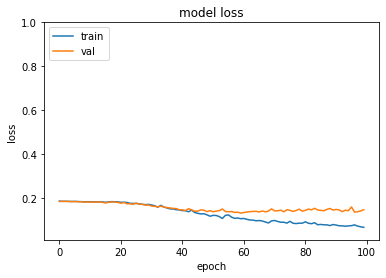}} \\ 
\caption{\label{figB}Evolution of Training and Validation Losses for alpha=1.0 (top), alpha=1.1 (middle) and alpha=1.3 (bottom) and 40 epochs}
\end{center}
\end{figure}

 We notice, thanks to figure \ref{figB}, that for $\alpha=1.1$,  $\alpha=1.3$, and $\alpha=1.0$, training and validation errors decrease at the same pace for N=40. Regarding Shannon's entropy, where the two losses are higher than for the two Havrda-Charvat-based models. We can deduce that, despite slightly decreasing too, Shannon's entropy is less relevant and trustworthy.\\
We can deduce that our results for 40 epochs regarding $\alpha=1.1$ are coherent with the tests. However, for a number of epochs N greater than 40, the phenomenon of overfitting starts to appear. We tested the models for N=100 epochs and the overfitting phenomenon appears and increases as the epochs happen. For the size of the studied dataset, 40 epochs can be considered as sufficient, but as the dataset's size increases, a greater number of epochs will be available without triggering any overfitting.\\
As a sum-up, for further study, the value $\alpha=1.1$ seems to be the most promising and an improvement over the commonly used Shannon's entropy.\\
Regarding the other value, we can notice that, after 40 to 50 epochs, the validation loss tends to increase, while the training loss reaches its low. If this value was to be used, one would have to consider that for a dataset of comparable size, epochs should be kept below 50, or adapted to reflect this level of fitting, since increasing epochs beyond this point would soon lower the adaptability of the network due to overfitting. \\

We notice that the validation error tends to increase far from the training one when N becomes superior to 30. This phenomenon tends to increase the further the epochs do so. Again, we can notice that epochs are not to be increased beyond 40, as the two losses seem to grow further away from one another at a rapid rate.

Moreover, we notice that the phenomenon becomes greater when using Shannon's entropy than with Havrda-Charvat's equation, which is another element in favor of the new loss function over the more commonly used Shannon-based loss. It is already starting at N=40 epochs for Shannon-based loss, as depicted in figure \ref{figB}, hence the lesser reliability of the results achieved thanks to it, including sensitivity and specificity.

\subsection{Images classification}

Among the correctly classified images, we find those with obvious, clearly identified structures that the network identifies as informative: 
\begin{figure}[h!]
\begin{center}
\subfigure{\includegraphics[width=0.1\textwidth]{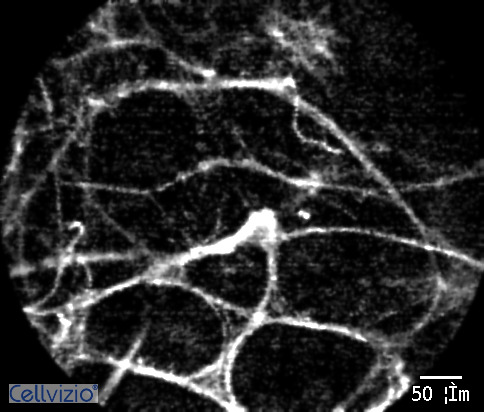}} \subfigure{\includegraphics[width=0.1\textwidth]{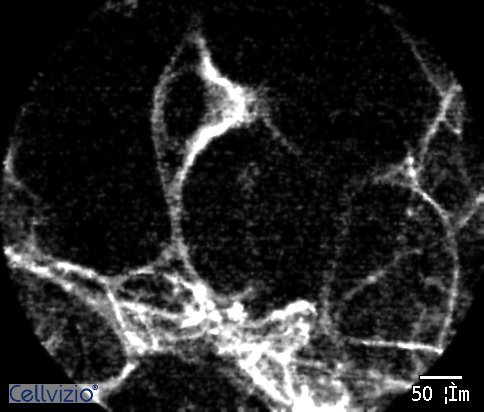}} \subfigure{\includegraphics[width=0.1\textwidth]{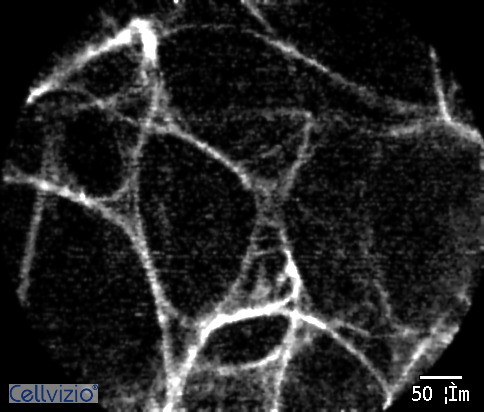}} \subfigure{\includegraphics[width=0.1\textwidth]{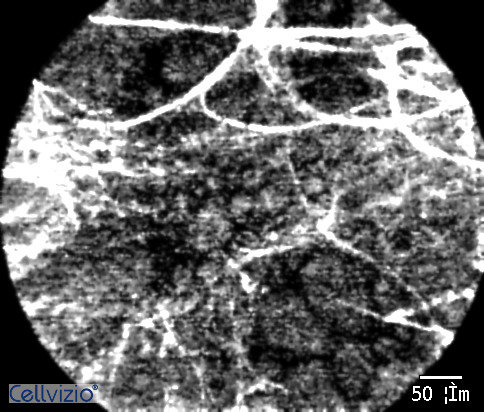}}\\
\caption{\label{figE} Several informative frames rightly classified}
\end{center}
\end{figure}

Conversely, the obviously uninformative ones are recognized by the algorithm and deemed as noise:

\begin{figure}[h!]
\begin{center}
\subfigure{\includegraphics[width=0.1\textwidth]{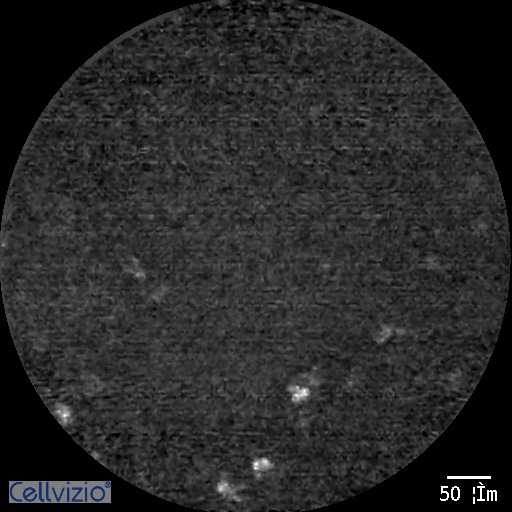}} \subfigure{\includegraphics[width=0.1\textwidth]{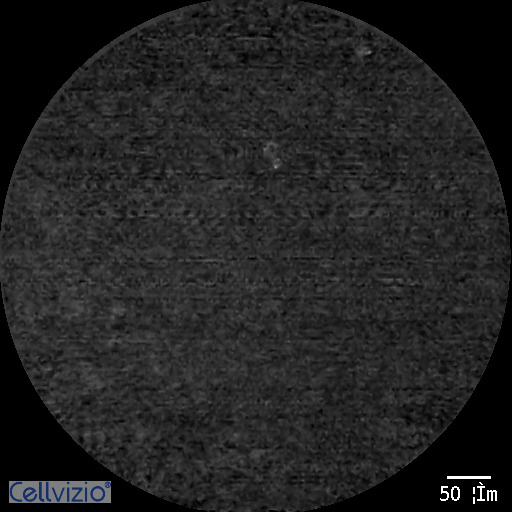}} \subfigure{\includegraphics[width=0.1\textwidth]{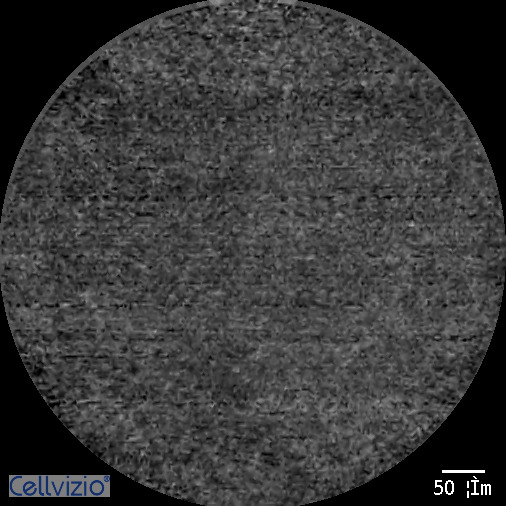}} \subfigure{\includegraphics[width=0.1\textwidth]{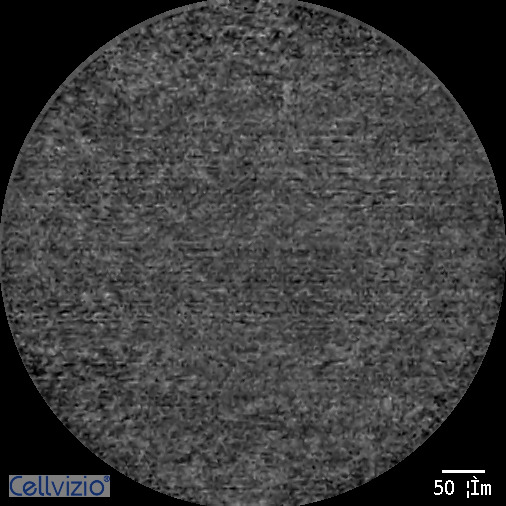}}\\
\caption{\label{figF} Several uninformative frames correctly classified}
\end{center}
\end{figure}

Despite the encouraging results achieved by the network, some limitations exist.

\section{Limitations}

\subsection{Classification errors}

During the experiments, we noticed that several images were classified in the wrong category. For example, the following images can be classified as informative although they were uninformative:

\begin{figure}[h!]
\begin{center}
\subfigure{\includegraphics[width=0.1\textwidth]{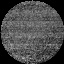}} \subfigure{\includegraphics[width=0.1\textwidth]{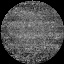}} \subfigure{\includegraphics[width=0.1\textwidth]{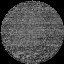}} \subfigure{\includegraphics[width=0.1\textwidth]{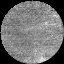}}\\
\caption{\label{figC} Several uninformative images classified as informative through the algorithm's error}
\end{center}
\end{figure}

As an explanation for the mistake, we can say that the data are very noisy and that any structure in the image can pass for an information and thus make the image an informative one, according to the neural network. Structures are hard to take in account, since the feature map at the end of the convolutional layers is flatten to meet the  fully connected layers' requirements for inputs.

Noise makes it difficult for the algorithm to notice when an artifact happens, when the patient moves for example or when the scope sees things that are not relevant to the current examination, like blood for example.

On the contrary, when the genuine structures become blurrier or are mixed with a lot of noise, which can result in the picture being deemed uninformative although a structure is present:

\begin{figure}[h!]
\begin{center}
\subfigure{\includegraphics[width=0.1\textwidth]{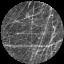}} \subfigure{\includegraphics[width=0.1\textwidth]{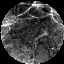}} \subfigure{\includegraphics[width=0.1\textwidth]{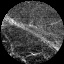}} \subfigure{\includegraphics[width=0.1\textwidth]{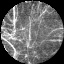}}\\
\caption{\label{figD} Several informative images classified as uninformative through the algorithm's error}
\end{center}
\end{figure}

So, noisy images still presenting informative elements are susceptible to being considered as being only noise and thus uninformative, the useful data being lost among the noise's values

A possible solution would be to pre-process the data with smoothing filters in order to reduce the noise's influence in the signal. 

\subsection{Data Scarcity and Overfitting}

As presented in the previous part, there's a phenomenon of overfitting the more we increase the number of epochs. This phenomenon comes from the small dataset that we use. Usually, the more epochs the better, but with smaller databases, the network tends to learn too well its training set, and become unable to adapt to any other data. This is why, for further research, we'll need more labeled data.

Another option regarding overfitting is to lower the complexity of the network by reducing the number of neurons in existing layers or by deleting some layers without altering the global structure of the network.

Regarding the labeled data scarcity, the solution is to create algorithms that are less impacted by small datasets, or that can work without labels on bigger sets of images.

\subsection{Balanced dataset}

In the foreword of this paper, we described the uninformative data as being about 25\% of the available frames of endomicroscopy videos. But, to make the learning less biased, we proposed to make each category (uninformative/informative)be closer to half the training dataset.\\ These data were chosen in order to be closer from a balanced dataset, but without cutting a sequence of frames. To make sure the dataset is completely representative of the real acquired data, further work should proceed with 25\% of the frames being uninformative to see if the results are similar. Unbalance in dataset is responsible for biases that make the algorithm determine that, since one class has a greater probability to occur than the other, then any analyzed frame will naturally be more prone to be categorized as such, regardless of its characteristics, as it's an issue of availability.
\subsection{Hyperparameter $\alpha$}
In this study, we proposed to set the parameter $\alpha$ to predetermined values, and have obtained the corresponding results. We noticed that, depending on the data available, the best value of the hyperparameter tended to vary. This is why we can assume that there may exist values of $\alpha$ yielding better results in the considered interval. In order to further improve our findings, the algorithm needs to be modified in order to automatically test the values of $\alpha$ and provide the value which gave the best results.

\section{Conclusion}
In this paper, we design a CNN classifier with a new loss function based on Havrda-Charvat entropy. Most of CNN classifier use Shannon entropy, while Havrda-Charvat entropy is a generalized Shannon entropy. Therefore, it can outperform it if the nature of data cannot satisfy certain conditions.  Our application aims to classify pulmonary optical endomicroscopy images in which informative images and non-informative images are not easy to distinguish. The proposed classifier can achieve an accuracy of 79\% (77\% for the second set), better than the one obtained by Shannon's entropy. In future work, we will analyze the endomicroscopy after removing the non-informative images for helping pathological diagnostic.

\section{Acknowledgment}

This project was co-financed by the European Union with the European  regional development fund (ERDF,\\
 18P03390/18E01750/18P02733) and by the Haute-Normandie Regional Council via the M2SINUM project.

\bibliographystyle{elsarticle-num}

\bibliography{strings}

\vspace{12pt}
\color{red}

\end{document}